\pdfoutput=1
\documentclass[letterpaper]{article}

  \usepackage{geometry}
  \usepackage[T1]{fontenc}
  \usepackage{newtxtext}
  \usepackage{helvet}
  \usepackage{courier}
  \usepackage{xcolor}
  \usepackage{natbib}
  \usepackage{caption}
  \usepackage{wrapfig}
  \usepackage{worldflags}
  \usepackage{needspace}
  \date{}
\makeatletter
\let\@dblfloat\@float
\let\end@dblfloat\end@float
\makeatother

\usepackage{amsmath,amsfonts,bm}

\def\eqref#1{equation~\ref{#1}}

\def\1{\bm{1}}

\DeclareMathAlphabet{\mathsfit}{\encodingdefault}{\sfdefault}{m}{sl}
\SetMathAlphabet{\mathsfit}{bold}{\encodingdefault}{\sfdefault}{bx}{n}

\usepackage{hyperref}
\usepackage{url}
\usepackage{xurl}
\usepackage{booktabs}
\usepackage{pgfplots}
\usepgfplotslibrary{fillbetween}
\usetikzlibrary{patterns,positioning,calc,arrows.meta}
\pgfplotsset{compat=1.18}
\usepackage{array}

\usepackage{xspace}
\usepackage{alltt}
\usepackage{cleveref}

\title{SqueezeBench: Saturation-Resilient Evaluation of Mathematical Capability}

\newcommand{\SB}{\textsc{SqueezeBench}\xspace}

\author{
  Pablo Manrique \qquad Hai Xia \qquad Stefan Szeider\\[4pt]
  \small Algorithms and Complexity Group\\[-2pt]
  \small TU Wien, Vienna, Austria\\[-2pt]
  \small \texttt{\{pmanriqu,hxia,sz\}@ac.tuwien.ac.at}
}
\hypersetup{pdftitle={SqueezeBench: Saturation-Resilient Evaluation of Mathematical Capability},%
  pdfauthor={Pablo Manrique, Hai Xia, Stefan Szeider}}
\hypersetup{hidelinks}
\begin{document}

\maketitle
\thispagestyle{empty}

\begin{abstract}
AI systems are making rapid progress in mathematics, but this progress is hard to measure. Benchmarks of solved problems saturate, and in benchmarks of open problems each problem can be resolved for the first time only once. We present \SB, a benchmark that is resilient to saturation: a success raises the bar instead of removing the problem. Each problem fixes an integer function $f(n)$ whose values are known only for small~$n$, and the benchmark keeps a lower and an upper record on it. A record is a computable Lean function with a machine-checked proof that it bounds~$f$. A submission proposes a new function and proves that it bounds~$f$, that it is no worse than the current record at any~$n$, and that it is strictly better at some~$n$. Once accepted, it becomes the record that the next submission has to beat. The system thus chooses the bound as well as its proof, and it may improve the record at a single~$n$ or improve its growth rate. At every~$n$, the gap between the two records shows how far the certified bounds are from determining $f(n)$. Every proof is checked by two independent Lean kernels, so no referee needs to check the mathematics. The checks also reject a function that computes~$f$ by exhaustive search at small~$n$, and a maintainer only reads the submitter's report and looks for such a search at larger~$n$. We release ten problems from extremal combinatorics with twenty proved initial records, together with the admission tools. As a baseline, agents based on a frontier language model, with at most one hour per record, had 82 submissions accepted and improved every record, without going beyond the published literature. Agents based on two newer models then started from these records, had 60 more submissions accepted and improved 19 of the 20 records again. For eight of the ten problems, at least one record is still asymptotically behind the best published bound.

\end{abstract}

\section{Introduction}

AI systems have made rapid progress in mathematics. On miniF2F, a benchmark of formalized olympiad problems, the best prover solved 29.2\% of the test problems when the benchmark appeared in 2021~\citep{zheng2022minif2f}, and 99.6\% four years later~\citep{chen2025seedprover}. In 2025, a system reached a gold-medal score at the International Mathematical Olympiad, certified by the competition's coordinators~\citep{luong2025imo}, and AI systems have since resolved a few open problems posed by Erd\H{o}s~\citep{feng2026erdos}.

This progress is hard to measure. A benchmark of solved problems has a maximum score, and as the best systems approach it, their scores can no longer be told apart~\citep{akhtar2026plateau}. Four systems have reported solving all 672 problems of PutnamBench~\citep{tsoukalas2024putnambench,putnambench2026leaderboard}, and the maintainers of FrontierMath consider its hardest tier saturated~\citep{epoch2026brief}. The answers of such benchmarks are often public, so they may also end up in the training data of the systems they evaluate. Benchmarks of open problems, such as FrontierMath Erd\H{o}s~\citep{adamczewski2026frontiermatherdos}, avoid known answers, but each problem yields at most one first resolution. After that, its solution is public, and a later system that solves it may be reproducing a published solution, so the set of informative problems shrinks as the systems improve. Their scores count resolved problems, and intermediate progress, such as a better bound, earns little or no credit. In both kinds of benchmark the statement to be proved is given, and the system is not asked to find it. Record archives such as EinsteinArena~\citep{bianchi2026einsteinarena} and Optimization Constants in Mathematics~\citep{davis2026optimizationconstants} keep improvable records, but a record there is a number for a fixed problem, checked by running a witness or by citing a paper, and no kernel checks a bound's proof.

We propose \SB, a benchmark designed to avoid these difficulties. Each problem fixes a function $f\colon\mathbb{N}\to\mathbb{N}$ whose values are known only for small~$n$, such as the size of the largest cap set in $\mathbb{F}_3^n$ or the diagonal Ramsey number $R(n,n)$. For each problem the benchmark keeps two records, a lower bound and an upper bound on~$f$. A record is a computable function written in Lean~4~\citep{moura2021lean4} together with a proof that it bounds~$f$. A submission proposes a new function and proves three obligations: it bounds~$f$, it is no worse than the current record at any~$n$, and it is strictly better at some~$n$. An accepted submission becomes the new record, and the next submission has to beat it, so a success leaves the problem in the benchmark with a stronger record. We call this property \emph{resilience}. It does not guarantee that systems remain distinguishable, since the benchmark stagnates when no system improves any record. At each~$n$ the two records enclose the unknown value $f(n)$, and the distance between them shows how far the certified bounds are from determining $f(n)$ (\Cref{fig:gap}). We include only problems with no known closed formula for~$f$.

Each accepted submission is stored with its author and date. At any time each record has a holder, and the record history of a problem shows who held which bound, when, and by how much each submission moved it. This history is the leaderboard. We do not reduce it to a single score, since progress on twenty records, each a function of~$n$, has many dimensions. To compare systems in a controlled way, we start each of them from the same snapshot of records with the same budget and compare the records they reach, value by value, not the number of accepted submissions. Replaying a bound that the current record already matches earns nothing, and a bound beyond the published literature cannot have been copied from it. Our initial records are elementary bounds weaker than the published ones, so the first steps can draw on the literature, and we report this catch-up phase as such.

The system proposes the bound itself, and only the three obligations are fixed. An improvement at a single~$n$ is accepted, and so is an improvement of the asymptotic growth. This freedom has a catch. The function~$f$ is itself a correct bound, and a record that computes~$f$ by exhaustive search could never be improved. The checks therefore evaluate each new record at every $n\le 100$ within a fixed budget, which rejects such a search below $n=100$, and a maintainer reads the submitter's report and looks for a search placed above it. Otherwise admission needs no mathematical referee. The proofs are verified by two independent Lean kernels against statements fixed before the submission, and the other checks also run mechanically. Once per problem, a maintainer also checks that the formal definition matches the intended question.

As a baseline, we ran agents based on a frontier language model, Opus~5 in Claude Code, one per record, with at most one hour each and 18.9 hours in total, starting from the initial records. They had 82 submissions accepted and improved every record at least once, for example the upper bound for the Ruzsa--Szemer\'edi problem from $\binom{n}{3}$ to $\lfloor n(n+3)/18 \rfloor$. Nearly all accepted submissions use general arguments and explicit constructions, not exhaustive search. None goes beyond the published literature, and for every problem at least one record stays asymptotically behind the literature.

We then ran agents based on two newer models from the baseline's records, Fable~5.1 on 19 records and Opus~5.5 on the lower record for Golomb rulers, in one run of up to about two hours per record, and the agents could consult a second language model. This continuation is not a controlled comparison of models but a test of resilience. The records left by one system were the starting point of the next, and the newer agents had 60 submissions accepted and improved 19 of the 20 records again. The number of records that match the leading term of the best published bound rose from four to ten, for example through the Ellenberg--Gijswijt bound for cap sets and Shearer's bound for $R(3,k)$, both formalized by the agent itself. All agents could search the web. One of them found a public Lean formalization of a published construction and adapted it, which the benchmark accepts as part of the catch-up phase when the license allows it and the source is credited. Neither run exhausted the problems. For eight of the ten problems, at least one record is still asymptotically behind the best published bound (\Cref{tab:problems}).

Our contributions are as follows.
\begin{itemize}
\item A benchmark design that measures mathematical progress without a fixed target: lower and upper records that are functions of the parameter, replaced only by submissions that prove the three obligations (\Cref{sec:framework}).
\item Ten problems from extremal combinatorics, formalized in Lean~4 on top of Mathlib, each with proved initial lower and upper records (\Cref{sec:problems}).
\item An admission pipeline that checks a submission mechanically against statements fixed in advance, with two independent kernels, the three standard axioms and a computation budget, together with tools to prepare, verify and accept submissions (\Cref{sec:framework}).
\item An evaluation of agents based on three frontier language models on all twenty records, a baseline from the initial records and a continuation from the baseline's records, with the harness that runs the baseline (\Cref{sec:evaluation}).
\end{itemize}
All problems, records and tools are publicly available at:

  \begin{center}
    \url{https://github.com/leansolving/SqueezeBench}
  \end{center}

\section{Framework}
\label{sec:framework}

As part of our benchmark, we formalize ten problems in the field of extremal combinatorics. This branch of mathematics studies how large or how small a finite object can be, subject to certain conditions. For instance, as we will further explain in~\Cref{sec:golomb}, \SB includes the \emph{Golomb ruler} problem~\citep{bloom1977golomb}, which asks the following question: for every natural number~$n$, what is the smallest possible length of a set of~$n$ integers whose pairwise differences are all distinct, where the length is the difference between the largest and the smallest integer? In other words, these are rulers with~$n$ marks that measure no distance twice. For small values of~$n$, one could even solve this problem by hand: for $n=3$, the ruler $\{0,1,3\}$ has length~3, but no ruler of length~2 exists, since $\{0,1,2\}$ measures the distance 1 twice. Thus, the smallest possible length is~3. For $n=4$, one can find rulers of length~6, such as $\{0,1,4,6\}$, but not of length~5, since the six pairwise differences would be distinct positive integers no larger than 5. Therefore, the shortest ruler has length~6. However, as the value of~$n$ increases, finding such shortest rulers becomes intractable, since there are $\binom{L-1}{n-2}$ candidate rulers of length~$L$. Although there is no known solution for the Golomb ruler problem for all possible values of~$n$, researchers have proved upper and lower bounds on these numbers~\citep{singer1938,erdos1941sidon,lindstrom1969}. This problem is typical of the problems in \SB. For each of them, the exact answer is known only for a small number of values of the parameter, or for special families of them, and no closed formula is known for the general case. In contrast, we consider bounds that are explicit: each has to be a computable function that can be evaluated at every value of the parameter. They can be closed formulas, tables or programs that build explicit constructions. The task is to narrow the gap between these known functions and the unknown answer. Note that our requirement on bounds is stronger than what the literature typically provides, where the best bounds are often asymptotic, with unspecified constants or valid only for sufficiently large values.

In short, \SB keeps the best lower and upper bounds on each~$f$ proved so far and accepts a submission only if it provably improves one of them. The rest of this section covers how a problem is formalized, what counts as a bound, how records are kept, what a submission consists of and how it is verified.

\paragraph{Problems.}
We have formalized each problem in Lean~4, building on its mathematical library, Mathlib~\citep{mathlib2020}. The answer to each problem is a function $f : \mathbb{N} \to \mathbb{N}$, where $f(n)$ is the optimal value for the parameter~$n$. In Lean, $f$ is defined by its combinatorial meaning, typically as a maximum or minimum over a finite set, but no closed form for~$f$ is known. This is what makes the benchmark resilient to saturation: if~$f$ had a closed form, the lower and upper bounds could meet at every~$n$, and the problem would be solved. This could eventually happen---and we would welcome a submission that settles an open problem---but we select problems for which it is not expected soon.
Each problem also fixes a smallest parameter value $n_{\min}$, in practice 1, 2 or 3, below which~$f$ is set to $0$. Some definitions degenerate at such small values, and pinning them lets a bound be stated for every~$n$ without side conditions.

\begin{figure}[t]
\centering
\definecolor{sbB250}{HTML}{86B6EF}
\definecolor{sbB300}{HTML}{6DA7EC}
\definecolor{sbB350}{HTML}{5598E7}
\definecolor{sbB400}{HTML}{3987E5}
\definecolor{sbB450}{HTML}{2A78D6}
\definecolor{sbB500}{HTML}{256ABF}
\definecolor{sbB550}{HTML}{1C5CAB}
\definecolor{sbB600}{HTML}{184F95}
\definecolor{sbB650}{HTML}{104281}
\definecolor{sbB700}{HTML}{0D366B}
\definecolor{sbO250}{HTML}{FA936E}
\definecolor{sbO300}{HTML}{F67B4D}
\definecolor{sbO350}{HTML}{EF642A}
\definecolor{sbO400}{HTML}{E05208}
\definecolor{sbO450}{HTML}{CB4801}
\definecolor{sbO500}{HTML}{B43F03}
\definecolor{sbO550}{HTML}{9E3601}
\definecolor{sbO600}{HTML}{892E02}
\definecolor{sbO650}{HTML}{752601}
\definecolor{sbO700}{HTML}{611E01}
\begin{tikzpicture}
\pgfplotsset{sbgolomb/.style={width=.52\textwidth, height=4.6cm, ymin=0,
  xlabel={number of marks~$n$}, tick label style={font=\scriptsize}, label style={font=\small},
  /pgf/number format/1000 sep={}}}
\begin{axis}[sbgolomb, name=ax1, xmin=1, xmax=100, ymax=9000, xtick={1,28,50,100},
  ytick={0,4000,8000}, ylabel={ruler length}, title={\small (a) lower records}]
  \addplot[semithick, sbB250, densely dashed] coordinates
    {(1,0)(4,6)(7,21)(10,45)(13,78)(16,120)(19,171)(22,231)(25,300)(28,378)(31,465)(34,561)(37,666)(40,780)(43,903)(46,1035)(49,1176)(52,1326)(55,1485)(58,1653)(61,1830)(64,2016)(67,2211)(70,2415)(73,2628)(76,2850)(79,3081)(82,3321)(85,3570)(88,3828)(91,4095)(94,4371)(97,4656)(100,4950)};
  \addplot[semithick, sbB350] coordinates
    {(1,0)(4,6)(7,22)(10,50)(13,93)(16,148)(19,221)(22,308)(25,407)(28,525)(31,658)(34,806)(37,966)(40,1147)(43,1343)(46,1554)(49,1777)(52,2022)(55,2282)(58,2558)(61,2850)(64,3153)(67,3479)(70,3820)(73,4178)(76,4552)(79,4941)(82,5343)(85,5768)(88,6208)(91,6665)(94,7137)(97,7626)(100,8127)};
  \addplot[semithick, sbB400] coordinates
    {(1,0)(4,6)(7,22)(10,51)(13,93)(16,150)(19,221)(22,308)(25,409)(28,525)(31,658)(34,806)(37,969)(40,1147)(43,1343)(46,1554)(49,1781)(52,2023)(55,2282)(58,2558)(61,2850)(64,3157)(67,3480)(70,3820)(73,4178)(76,4552)(79,4941)(82,5347)(85,5769)(88,6208)(91,6665)(94,7137)(97,7626)(100,8132)};
  \addplot[semithick, sbB450] coordinates
    {(1,0)(4,6)(7,22)(10,51)(13,94)(16,151)(19,222)(22,309)(25,411)(28,527)(31,660)(34,808)(37,971)(40,1150)(43,1346)(46,1557)(49,1784)(52,2026)(55,2286)(58,2562)(61,2854)(64,3161)(67,3485)(70,3826)(73,4184)(76,4557)(79,4947)(82,5353)(85,5774)(88,6215)(91,6671)(94,7144)(97,7633)(100,8138)};
  \addplot[semithick, sbB500] coordinates
    {(1,0)(4,6)(7,22)(10,51)(13,95)(16,152)(19,223)(22,310)(25,411)(28,528)(31,661)(34,809)(37,972)(40,1152)(43,1347)(46,1559)(49,1785)(52,2028)(55,2288)(58,2564)(61,2856)(64,3163)(67,3487)(70,3828)(73,4186)(76,4560)(79,4949)(82,5355)(85,5777)(88,6218)(91,6674)(94,7147)(97,7636)(100,8141)};
  \addplot[semithick, sbB550] coordinates
    {(1,0)(4,6)(7,22)(10,51)(13,95)(16,152)(19,224)(22,311)(25,412)(28,529)(31,662)(34,810)(37,973)(40,1152)(43,1348)(46,1560)(49,1786)(52,2029)(55,2289)(58,2565)(61,2857)(64,3164)(67,3488)(70,3830)(73,4187)(76,4561)(79,4951)(82,5356)(85,5779)(88,6219)(91,6676)(94,7149)(97,7638)(100,8143)};
  \addplot[semithick, sbB600] coordinates
    {(1,0)(4,6)(7,23)(10,52)(13,96)(16,153)(19,225)(22,312)(25,413)(28,530)(31,663)(34,811)(37,974)(40,1154)(43,1350)(46,1562)(49,1788)(52,2032)(55,2292)(58,2568)(61,2860)(64,3167)(67,3492)(70,3834)(73,4191)(76,4565)(79,4955)(82,5360)(85,5784)(88,6224)(91,6681)(94,7154)(97,7643)(100,8148)};
  \addplot[semithick, sbB650] coordinates
    {(1,0)(4,6)(7,23)(10,53)(13,97)(16,154)(19,227)(22,314)(25,416)(28,534)(31,667)(34,815)(37,980)(40,1160)(43,1356)(46,1567)(49,1795)(52,2040)(55,2300)(58,2576)(61,2868)(64,3177)(67,3503)(70,3844)(73,4202)(76,4576)(79,4966)(82,5374)(85,5798)(88,6239)(91,6695)(94,7168)(97,7657)(100,8165)};
  \addplot[very thick, sbB700] coordinates
    {(1,0)(4,6)(7,23)(10,53)(13,97)(16,156)(19,229)(22,316)(25,419)(28,536)(31,668)(34,815)(37,987)(40,1168)(43,1363)(46,1574)(49,1808)(52,2053)(55,2314)(58,2591)(61,2884)(64,3197)(67,3524)(70,3868)(73,4228)(76,4604)(79,4996)(82,5407)(85,5832)(88,6274)(91,6732)(94,7206)(97,7696)(100,8210)};
  \node[font=\scriptsize, text=black!70, anchor=west] at (axis cs:78,3000) {V000};
  \node[font=\scriptsize, text=black!70, anchor=east] at (axis cs:88,7400) {V001--V008};
\end{axis}
\begin{axis}[at={(ax1.north west)}, anchor=north west, xshift=0.4cm, yshift=-0.12cm,
  width=.29\textwidth, height=2.9cm, xmin=1, xmax=100, ymin=0, ymax=100, xtick={1,50,100},
  ytick={0,40,80}, tick label style={font=\tiny}, axis background/.style={fill=white}]
    \addplot[sbB400] coordinates
      {(1,0)(2,0)(3,0)(4,0)(5,0)(6,0)(7,0)(8,0)(9,2)(10,1)(11,0)(12,0)(13,0)(14,0)(15,0)(16,2)(17,1)(18,1)(19,0)(20,0)(21,0)(22,0)(23,0)(24,0)(25,2)(26,2)(27,1)(28,0)(29,0)(30,0)(31,0)(32,0)(33,0)(34,0)(35,0)(36,3)(37,3)(38,1)(39,1)(40,0)(41,0)(42,0)(43,0)(44,0)(45,0)(46,0)(47,0)(48,0)(49,4)(50,3)(51,2)(52,1)(53,0)(54,0)(55,0)(56,0)(57,0)(58,0)(59,0)(60,0)(61,0)(62,0)(63,0)(64,4)(65,3)(66,3)(67,1)(68,1)(69,0)(70,0)(71,0)(72,0)(73,0)(74,0)(75,0)(76,0)(77,0)(78,0)(79,0)(80,0)(81,4)(82,4)(83,3)(84,2)(85,1)(86,0)(87,0)(88,0)(89,0)(90,0)(91,0)(92,0)(93,0)(94,0)(95,0)(96,0)(97,0)(98,0)(99,0)(100,5)};
    \addplot[sbB450] coordinates
      {(1,0)(2,0)(3,0)(4,0)(5,0)(6,0)(7,0)(8,1)(9,2)(10,1)(11,1)(12,1)(13,1)(14,1)(15,1)(16,3)(17,2)(18,2)(19,1)(20,2)(21,2)(22,1)(23,2)(24,1)(25,4)(26,4)(27,2)(28,2)(29,3)(30,2)(31,2)(32,3)(33,2)(34,2)(35,3)(36,5)(37,5)(38,4)(39,3)(40,3)(41,4)(42,3)(43,3)(44,4)(45,3)(46,3)(47,3)(48,4)(49,7)(50,6)(51,6)(52,4)(53,4)(54,5)(55,4)(56,4)(57,4)(58,4)(59,5)(60,4)(61,4)(62,5)(63,4)(64,8)(65,7)(66,7)(67,6)(68,5)(69,5)(70,6)(71,5)(72,6)(73,6)(74,5)(75,6)(76,5)(77,5)(78,5)(79,6)(80,5)(81,10)(82,10)(83,8)(84,8)(85,6)(86,6)(87,7)(88,7)(89,7)(90,6)(91,6)(92,7)(93,7)(94,7)(95,6)(96,6)(97,7)(98,7)(99,7)(100,11)};
    \addplot[sbB500] coordinates
      {(1,0)(2,0)(3,0)(4,0)(5,0)(6,0)(7,0)(8,1)(9,2)(10,1)(11,1)(12,2)(13,2)(14,1)(15,1)(16,4)(17,3)(18,2)(19,2)(20,2)(21,2)(22,2)(23,2)(24,2)(25,4)(26,4)(27,3)(28,3)(29,4)(30,3)(31,3)(32,4)(33,3)(34,3)(35,4)(36,6)(37,6)(38,5)(39,5)(40,5)(41,5)(42,4)(43,4)(44,5)(45,5)(46,5)(47,5)(48,5)(49,8)(50,7)(51,7)(52,6)(53,6)(54,6)(55,6)(56,6)(57,6)(58,6)(59,6)(60,6)(61,6)(62,6)(63,6)(64,10)(65,9)(66,9)(67,8)(68,8)(69,8)(70,8)(71,8)(72,8)(73,8)(74,8)(75,8)(76,8)(77,8)(78,8)(79,8)(80,8)(81,12)(82,12)(83,11)(84,10)(85,9)(86,9)(87,10)(88,10)(89,10)(90,9)(91,9)(92,10)(93,10)(94,10)(95,9)(96,9)(97,10)(98,10)(99,10)(100,14)};
    \addplot[sbB550] coordinates
      {(1,0)(2,0)(3,0)(4,0)(5,0)(6,0)(7,0)(8,1)(9,2)(10,1)(11,1)(12,2)(13,2)(14,1)(15,1)(16,4)(17,3)(18,3)(19,3)(20,3)(21,3)(22,3)(23,3)(24,3)(25,5)(26,5)(27,4)(28,4)(29,4)(30,4)(31,4)(32,4)(33,4)(34,4)(35,4)(36,7)(37,7)(38,5)(39,6)(40,5)(41,6)(42,5)(43,5)(44,6)(45,5)(46,6)(47,5)(48,6)(49,9)(50,8)(51,8)(52,7)(53,7)(54,8)(55,7)(56,7)(57,7)(58,7)(59,8)(60,7)(61,7)(62,8)(63,7)(64,11)(65,10)(66,10)(67,9)(68,9)(69,9)(70,10)(71,9)(72,9)(73,9)(74,9)(75,10)(76,9)(77,9)(78,9)(79,10)(80,9)(81,13)(82,13)(83,12)(84,12)(85,11)(86,11)(87,12)(88,11)(89,12)(90,11)(91,11)(92,12)(93,11)(94,12)(95,11)(96,11)(97,12)(98,11)(99,12)(100,16)};
    \addplot[sbB600] coordinates
      {(1,0)(2,0)(3,0)(4,0)(5,0)(6,1)(7,1)(8,1)(9,3)(10,2)(11,2)(12,3)(13,3)(14,2)(15,2)(16,5)(17,4)(18,4)(19,4)(20,4)(21,4)(22,4)(23,4)(24,4)(25,6)(26,6)(27,5)(28,5)(29,6)(30,5)(31,5)(32,6)(33,5)(34,5)(35,6)(36,8)(37,8)(38,7)(39,8)(40,7)(41,8)(42,7)(43,7)(44,8)(45,7)(46,8)(47,7)(48,8)(49,11)(50,10)(51,11)(52,10)(53,10)(54,11)(55,10)(56,10)(57,10)(58,10)(59,11)(60,10)(61,10)(62,11)(63,10)(64,14)(65,13)(66,14)(67,13)(68,13)(69,13)(70,14)(71,13)(72,13)(73,13)(74,13)(75,14)(76,13)(77,13)(78,13)(79,14)(80,13)(81,17)(82,17)(83,16)(84,17)(85,16)(86,16)(87,17)(88,16)(89,17)(90,16)(91,16)(92,17)(93,16)(94,17)(95,16)(96,16)(97,17)(98,16)(99,17)(100,21)};
    \addplot[sbB650] coordinates
      {(1,0)(2,0)(3,0)(4,0)(5,1)(6,1)(7,1)(8,2)(9,4)(10,3)(11,3)(12,4)(13,4)(14,3)(15,3)(16,6)(17,6)(18,6)(19,6)(20,6)(21,6)(22,6)(23,6)(24,6)(25,9)(26,10)(27,9)(28,9)(29,10)(30,9)(31,9)(32,10)(33,9)(34,9)(35,10)(36,13)(37,14)(38,13)(39,13)(40,13)(41,14)(42,13)(43,13)(44,14)(45,13)(46,13)(47,13)(48,14)(49,18)(50,18)(51,19)(52,18)(53,18)(54,19)(55,18)(56,18)(57,18)(58,18)(59,19)(60,18)(61,18)(62,19)(63,19)(64,24)(65,24)(66,24)(67,24)(68,24)(69,24)(70,24)(71,24)(72,24)(73,24)(74,24)(75,24)(76,24)(77,24)(78,24)(79,25)(80,25)(81,30)(82,31)(83,31)(84,31)(85,30)(86,30)(87,31)(88,31)(89,31)(90,30)(91,30)(92,31)(93,31)(94,31)(95,30)(96,30)(97,31)(98,32)(99,33)(100,38)};
    \addplot[thick, sbB700] coordinates
      {(1,0)(2,0)(3,0)(4,0)(5,1)(6,1)(7,1)(8,2)(9,4)(10,3)(11,3)(12,4)(13,4)(14,3)(15,3)(16,8)(17,8)(18,8)(19,8)(20,8)(21,8)(22,8)(23,8)(24,8)(25,12)(26,13)(27,12)(28,11)(29,12)(30,11)(31,10)(32,10)(33,9)(34,9)(35,10)(36,20)(37,21)(38,21)(39,21)(40,21)(41,21)(42,20)(43,20)(44,21)(45,20)(46,20)(47,20)(48,21)(49,31)(50,31)(51,32)(52,31)(53,32)(54,33)(55,32)(56,33)(57,33)(58,33)(59,34)(60,34)(61,34)(62,35)(63,35)(64,44)(65,44)(66,45)(67,45)(68,46)(69,46)(70,48)(71,48)(72,49)(73,50)(74,50)(75,52)(76,52)(77,53)(78,53)(79,55)(80,55)(81,63)(82,64)(83,64)(84,65)(85,64)(86,65)(87,66)(88,66)(89,67)(90,66)(91,67)(92,68)(93,68)(94,69)(95,68)(96,69)(97,70)(98,70)(99,71)(100,83)};
  \node[font=\tiny, anchor=north west] at (axis cs:3,97) {gain $l_k(n)-l_1(n)$};
  \node[font=\tiny, text=black!70, anchor=east] at (axis cs:52,58) {V008};
\end{axis}
\begin{axis}[sbgolomb, at={(ax1.east)}, anchor=west, xshift=0.9cm, xmin=1, xmax=40, ymax=1400,
  xtick={1,10,20,28,40}, ytick={0,500,1000}, title={\small (b) upper records}]
  \addplot[semithick, sbO250, densely dashed] coordinates
    {(1,0)(2,1)(3,3)(4,7)(5,11)(6,17)(7,63)(8,127)(9,255)(10,511)(11,1023)(12,2047)};
  \addplot[semithick, sbO350] coordinates
    {(1,0)(2,1)(3,3)(4,7)(5,11)(6,17)(7,63)(8,127)(9,255)(10,379)(11,461)(12,551)(13,649)(14,755)(15,869)(16,991)(17,1121)(18,1259)(19,1405)};
  \addplot[semithick, sbO450] coordinates
    {(1,0)(2,1)(3,3)(4,7)(5,11)(6,17)(7,63)(8,127)(9,185)(10,202)(11,269)(12,290)(13,416)(14,458)(15,485)(16,514)(17,617)(18,650)(19,830)(20,890)(21,929)(22,970)(23,1296)(24,1341)(25,1417)};
  \addplot[semithick, sbO500] coordinates
    {(1,0)(2,1)(3,3)(4,6)(5,11)(6,17)(7,25)(8,34)(9,44)(10,55)(11,72)(12,290)(13,416)(14,458)(15,485)(16,514)(17,617)(18,650)(19,830)(20,890)(21,929)(22,970)(23,1296)(24,1341)(25,1417)};
  \addplot[semithick, sbO500] coordinates
    {(1,0)(2,1)(3,3)(4,6)(5,11)(6,17)(7,25)(8,34)(9,44)(10,55)(11,72)(12,85)(13,106)(14,127)(15,151)(16,514)(17,617)(18,650)(19,830)(20,890)(21,929)(22,970)(23,1296)(24,1341)(25,1417)};
  \addplot[semithick, sbO500] coordinates
    {(1,0)(2,1)(3,3)(4,6)(5,11)(6,17)(7,25)(8,34)(9,44)(10,55)(11,72)(12,85)(13,106)(14,127)(15,151)(16,177)(17,199)(18,650)(19,830)(20,890)(21,929)(22,970)(23,1296)(24,1341)(25,1417)};
  \addplot[semithick, sbO550] coordinates
    {(1,0)(2,1)(3,3)(4,6)(5,11)(6,17)(7,25)(8,34)(9,44)(10,55)(11,72)(12,85)(13,106)(14,127)(15,151)(16,177)(17,199)(18,341)(19,505)(20,505)(21,505)(22,505)(23,811)(24,811)(25,811)(26,811)(27,811)(28,811)(29,929)(30,929)(31,1331)(32,1331)(33,1331)(34,1331)(35,1331)(36,1331)(37,1639)};
  \addplot[semithick, sbO600] coordinates
    {(1,0)(2,1)(3,3)(4,6)(5,11)(6,17)(7,25)(8,34)(9,44)(10,55)(11,72)(12,85)(13,106)(14,127)(15,151)(16,177)(17,199)(18,216)(19,246)(20,283)(21,333)(22,356)(23,372)(24,425)(25,480)(26,492)(27,553)(28,585)(29,929)(30,929)(31,1331)(32,1331)(33,1331)(34,1331)(35,1331)(36,1331)(37,1639)};
  \addplot[semithick, sbO600] coordinates
    {(1,0)(2,1)(3,3)(4,6)(5,11)(6,17)(7,25)(8,34)(9,44)(10,55)(11,72)(12,85)(13,106)(14,127)(15,151)(16,177)(17,199)(18,216)(19,246)(20,283)(21,333)(22,356)(23,372)(24,425)(25,480)(26,492)(27,553)(28,585)(29,868)(30,899)(31,1110)(32,1147)(33,1184)(34,1221)(35,1258)(36,1295)(37,1476)};
  \addplot[semithick, sbO600] coordinates
    {(1,0)(2,1)(3,3)(4,6)(5,11)(6,17)(7,25)(8,34)(9,44)(10,55)(11,72)(12,85)(13,106)(14,127)(15,151)(16,177)(17,199)(18,216)(19,246)(20,283)(21,333)(22,356)(23,372)(24,425)(25,480)(26,492)(27,553)(28,585)(29,811)(30,898)(31,929)(32,1146)(33,1183)(34,1220)(35,1257)(36,1294)(37,1331)(38,1516)};
  \addplot[semithick, sbO650] coordinates
    {(1,0)(2,1)(3,3)(4,6)(5,11)(6,17)(7,25)(8,34)(9,44)(10,55)(11,72)(12,85)(13,106)(14,127)(15,151)(16,177)(17,199)(18,216)(19,246)(20,283)(21,333)(22,356)(23,372)(24,425)(25,480)(26,492)(27,553)(28,585)(29,811)(30,841)(31,929)(32,961)(33,1183)(34,1220)(35,1257)(36,1294)(37,1331)(38,1369)(39,1557)};
  \addplot[semithick, sbO650] coordinates
    {(1,0)(2,1)(3,3)(4,6)(5,11)(6,17)(7,25)(8,34)(9,44)(10,55)(11,72)(12,85)(13,106)(14,127)(15,151)(16,177)(17,199)(18,216)(19,246)(20,283)(21,333)(22,356)(23,372)(24,425)(25,480)(26,492)(27,553)(28,585)(29,811)(30,841)(31,929)(32,961)(33,1024)(34,1220)(35,1257)(36,1294)(37,1331)(38,1369)(39,1557)};
  \addplot[semithick, sbO650] coordinates
    {(1,0)(2,1)(3,3)(4,6)(5,11)(6,17)(7,25)(8,34)(9,44)(10,55)(11,72)(12,85)(13,106)(14,127)(15,151)(16,177)(17,199)(18,216)(19,246)(20,283)(21,333)(22,356)(23,372)(24,425)(25,480)(26,492)(27,553)(28,585)(29,623)(30,680)(31,747)(32,784)(33,859)(34,938)(35,987)(36,1005)(37,1099)(38,1146)(39,1252)(40,1282)};
  \addplot[very thick, sbO700] coordinates
    {(1,0)(2,1)(3,3)(4,6)(5,11)(6,17)(7,25)(8,34)(9,44)(10,55)(11,72)(12,85)(13,106)(14,127)(15,151)(16,177)(17,199)(18,216)(19,246)(20,283)(21,333)(22,356)(23,372)(24,425)(25,480)(26,492)(27,553)(28,585)(29,623)(30,680)(31,747)(32,784)(33,859)(34,938)(35,987)(36,1005)(37,1099)(38,1146)(39,1252)(40,1282)};
  \node[font=\scriptsize, text=black!70, anchor=east] at (axis cs:9.6,1250) {V000};
  \node[font=\scriptsize, text=black!70, anchor=west] at (axis cs:14.5,1250) {V001};
  \node[font=\scriptsize, text=black!70, anchor=west] at (axis cs:20.5,1080) {V002};
  \node[font=\scriptsize, text=black!70, anchor=west] at (axis cs:21.5,700) {V006};
  \node[font=\scriptsize, text=black!70, anchor=west] at (axis cs:31.5,620) {V013};
\end{axis}
\end{tikzpicture}
\caption{Every accepted record for the Golomb ruler problem. Colors run from light for the oldest version to dark for the newest, the standing record is drawn thick, and each version is at least as good as its predecessor at every~$n$, which its submission proved. \emph{(a)} The lower records for $n\le 100$: the seed $\binom{n}{2}$ (dashed) and versions V001--V008, which differ only in lower-order terms and overlap at this scale. Writing $l_k$ for version V00$k$ of the lower record, the inset shows the gain $l_k(n)-l_1(n)$ of each later version over V001, for $k=2,\dots,8$ ($k=8$ drawn thick). \emph{(b)} The upper records for $n\le 40$, from the seed (dashed) to the standing record V013.}
\label{fig:gap}
\end{figure}
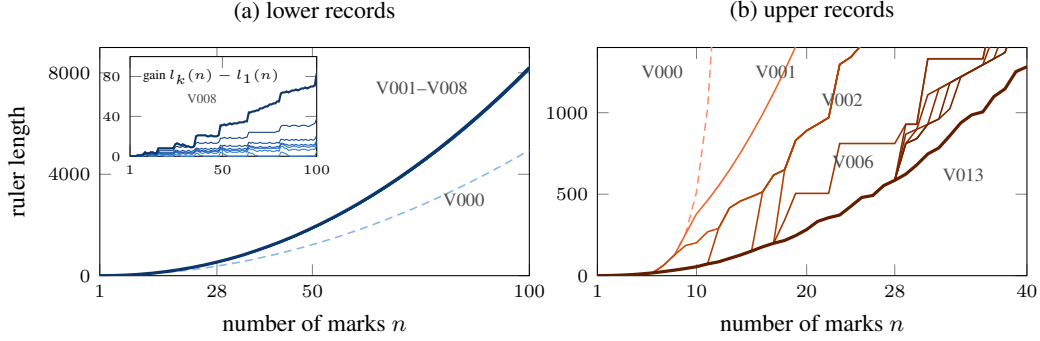

\paragraph{Bounds and records.}
A \emph{lower bound} on~$f$ is a function $l : \mathbb{N} \to \mathbb{N}$ with $l(n) \leq f(n)$ for all~$n$, and an \emph{upper bound} is a function~$u$ with $f(n) \leq u(n)$ for all~$n$. For each problem, \SB keeps a lower and an upper \emph{record}, the best bound proved so far in each direction, as a sequence of versions that starts from an elementary seed (\Cref{fig:gap}). The \emph{record gap} $u(n) - l(n)$ measures how far the records are from determining $f(n)$; for Golomb rulers, $536 \le f(28) \le 585$. A submission that replaces a lower record~$l$ consists of a new Lean function $l'$ and proofs of three theorems: \texttt{valid} ($l' \leq f$), \texttt{improves} ($l \leq l'$), both for all~$n$, and \texttt{strict} ($l(n_0) < l'(n_0)$ for some $n_0 \geq n_{\min}$). For upper records, the inequalities are reversed.
The \texttt{valid} theorem is usually the one that requires most of the proof effort in a submission. The other two are more easily automatable: we provide \emph{idioms} that allow defining a new bound from the current record by improving it at a single point or taking the pointwise maximum with another valid bound, so the \texttt{improves} requirement is trivial in these cases. For proving \texttt{strict}, the bounds are evaluated and compared at a witness point $n_0$.

\begin{figure}[t]
\centering
\begin{tikzpicture}[font=\footnotesize,
  sbtrusted/.style={draw, fill=gray!15, rounded corners=2pt, align=left, inner sep=4pt},
  sbuntrusted/.style={draw, dashed, rounded corners=2pt, align=left, inner sep=4pt},
  sbtool/.style={-{Stealth[length=2mm]}, thick, blue!55!black},
  sbimp/.style={-{Stealth[length=2mm]}, thick, densely dotted},
  sblab/.style={font=\scriptsize, text=blue!55!black}]
  \node[sbtrusted, minimum width=3.0cm] (ck) at (0,0) {%
    \textbf{Challenge $V_k$}\\[2pt]
    \scriptsize\texttt{import} $V_{k-1}$\texttt{/Def.lean}\\
    \scriptsize\texttt{record} \hfill open\\
    \scriptsize\texttt{valid} \hfill\texttt{sorry}\\
    \scriptsize\texttt{improves} \hfill\texttt{sorry}\\
    \scriptsize\texttt{strict} \hfill\texttt{sorry}};
  \node[sbuntrusted, right=1.75cm of ck] (sub) {%
    \textbf{Submission $V_k$}\\[2pt]
    \scriptsize\begin{tabular}{@{}l@{\hspace{5pt}}l@{}}
      \texttt{Def.lean} & the new record\\
      \texttt{Proof.lean} & \texttt{valid}, \texttt{improves}, \texttt{strict}\\
      \texttt{metadata.toml} & author, witness $n_0$\\
      \texttt{REPORT.md} & how the bound was found
    \end{tabular}};
  \node[draw, rounded corners=2pt, align=left, inner sep=4pt, right=0.7cm of sub] (ver) {%
    \textbf{\texttt{squeeze verify}}\\[2pt]
    \scriptsize 1 only its own directory\\
    \scriptsize 2 plain definitions\\
    \scriptsize 3 Comparator, 2 kernels\\
    \scriptsize 4 proof budget\\
    \scriptsize 5 computable, no~$f$\\
    \scriptsize 6 cheap for $n\le 100$};
  \node[draw, rounded corners=2pt, align=center, below=1.1cm of ver] (rev) {maintainer review};
  \node[sbtrusted, minimum width=3.0cm] (ck1) at (ck |- rev) {%
    \textbf{Challenge $V_{k+1}$}\\[2pt]
    \scriptsize\texttt{import} $V_{k}$\texttt{/Def.lean}};
  \draw[sbtool] (ck) -- node[sblab, above, align=center] {\texttt{squeeze}\\\texttt{new}} (sub);
  \draw[sbtool] (sub) -- node[sblab, above] {PR} (ver);
  \draw[sbtool] (ver) -- node[sblab, right] {all pass} (rev);
  \draw[sbtool] (rev) -- node[sblab, below] {\texttt{squeeze accept}} (ck1);
  \draw[sbimp] ($(sub.south west)+(0.9,0)$) to[out=-100, in=0] node[pos=0.22, font=\scriptsize, right, xshift=2pt] {\texttt{Def.lean} becomes trusted} ($(ck1.north east)+(0,-0.3)$);
\end{tikzpicture}
\caption{One version step: trusted files written by \texttt{squeeze} (solid), untrusted submission (dashed).}
\label{fig:lifecycle}
\end{figure}
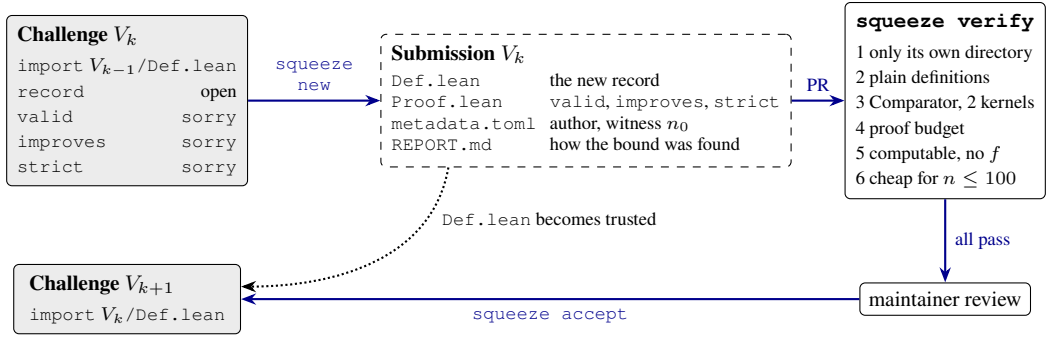

\paragraph{Challenges and submissions.}
A submission must be checked without trusting the person who wrote it. For this, we use Comparator~\citep{comparator}, a tool from the Lean FRO for judging Lean proofs, which is also used by other projects such as \texttt{lean-eval}~\citep{leaneval} and Palomar~\citep{palomar}. Comparator takes a trusted \emph{challenge}, which states theorems with their proofs left as \texttt{sorry} and may leave some definitions open, and an untrusted \emph{solution}, which fills them in. It checks that the solution proves exactly the statements of the challenge and uses no axioms beyond a given list. We allow only the three standard axioms \texttt{propext}, \texttt{Quot.sound} and \texttt{Classical.choice}, so \texttt{sorry} and \texttt{native\_decide} are rejected. It then checks the proofs with the Lean kernel and, in our setup, also with nanoda~\citep{nanoda}, an independent implementation of the kernel. Our design follows this interface. For every new version of a record, \SB provides a \emph{Challenge}: a trusted Lean file in which the new \texttt{record} definition is left open, and \texttt{valid}, \texttt{improves} and \texttt{strict} are stated in terms of~$f$ and the current record. The solution is split into two Lean files: \texttt{Def.lean} defines the new function, and \texttt{Proof.lean} imports it and proves the three theorems. The split matters because the next Challenge imports \texttt{Def.lean} to refer to this record, which makes that file trusted; keeping the proofs out of it keeps the trusted part small. Two more files complete the submission: \texttt{metadata.toml} gives the author name and the witness $n_0$, and \texttt{REPORT.md} explains how the bound and its proofs were obtained and the setup used, including the models or agents. \SB is hosted in a public GitHub repository, and a submission is a pull request that adds these four files.

\paragraph{Verification and acceptance.}
Every step is run by \texttt{squeeze}, a command-line tool that we implemented in Lean (\Cref{fig:lifecycle}). A submitter runs \texttt{squeeze new}, which creates the four files of a submission with the statements copied from the open Challenge, and opens a pull request that adds them. \texttt{squeeze verify} checks the submission in six \emph{gates}, both locally and on the pull request.
\begin{itemize}\itemsep0pt
\item \emph{Gates 1--2.} They run before any submitted code is built. The pull request may change only files inside the new record's directory, and \texttt{Def.lean} must be at most 64~KiB and contain only plain definitions, since later Challenges import it and anything else could alter them.
\item \emph{Gate 3.} Comparator, as described above.
\item \emph{Gates 4--6.} They check what Comparator does not. (4) The proofs in \texttt{Proof.lean} stay within a computation budget, per declaration and in total, so that they argue instead of making the kernel search by brute force. (5) The new record is computable, and its definition does not refer to~$f$. (6) The kernel evaluates the new record at every~$n$ from $n_{\min}$ to $100$, and at $n_0$ if it is larger, each within a fixed budget.
\end{itemize}
Gate 6 lets the next submitter evaluate the record to beat it. It also rejects a record that computes~$f$ by exhaustive search, such as a brute-force computation of~$f$ under another name, which passes every other gate; such a record is a correct bound that no later submission could improve. No known algorithm computes~$f(n)$ within the budget for $n$ near 100, while every record accepted so far passes. Budgets are counted in heartbeats, Lean's measure of work, so for fixed Lean and Mathlib versions they do not depend on the machine. Once all gates pass:
\begin{itemize}\itemsep0pt
\item \emph{Review.} A maintainer reads the submission, including its report, and merges it. The maintainer also rejects a search that starts above $n = 100$.
\item \emph{Next Challenge.} \texttt{squeeze accept} writes the next Challenge, stated against the new record. If two submissions answer the same Challenge, the first one merged wins, and \texttt{squeeze verify} rejects the other until it is restated against the new record.
\item \emph{Audit and history.} Since Challenges are never edited, \texttt{squeeze audit} can re-check every past version against the Challenge that admitted it. Each problem's page in the repository lists every accepted version with its date, author and witness.
\end{itemize}

\section{Related Work}

\SB lies between mathematical reasoning benchmarks, archives of best known results and formal theorem proving, and \Cref{tab:related} compares it with benchmarks and archives.

\begin{table}[t]
\centering
\caption{Mathematical benchmarks and record archives. \emph{NA} marks a system that
maintains no standing record. Only where a submission replaces a \emph{total function} can a replacement be tighter at one~$n$ and looser at another, and only \SB both replaces functions and proves that the replacement never regresses.
\textbf{Notes:} $^{a}$~Formal Conjectures' function-valued bound goals are unproved
(\texttt{:= by sorry}) and asymptotic, so improving a single~$n$ earns no credit.
$^{b}$~HorizonMath compares against a fixed value in \texttt{data/baselines.json}; beating it does
not replace it. $^{c}$~SorryDB and ArXivLean refresh the task stream: an answer key exists, but is kept
fresh. $^{d}$~Formal Conjectures' anchors \texttt{c1a\_lower\_bound} and \texttt{c1a\_upper\_bound}
both carry \texttt{sorry} while tagged as solved. $^{e}$~FrontierShift shows a gap on some pages,
but only its upper side is a maintained record.
\textbf{Sources.} PutnamBench \citep{tsoukalas2024putnambench}, FormalMATH \citep{formalmath2025},
Lean Workbook \citep{ying2024leanworkbook}, CombiBench \citep{liu2025combibench}, SorryDB
\citep{letson2026sorrydb}, ArXivLean \citep{matharena2026arxivlean}, FM: Open Problems~\citep{epoch2026frontiermathopen}, Covering \citep{coveringrepository2026}, Grassl
\citep{grassl2026codetables}, EinsteinArena \citep{bianchi2026einsteinarena}. The remaining
systems are cited above.}
\label{tab:related}
\footnotesize
\setlength{\tabcolsep}{1.5pt}
\renewcommand{\arraystretch}{1.1}
\hyphenpenalty=10000 \exhyphenpenalty=10000
\newcolumntype{R}[1]{>{\raggedright\arraybackslash}p{#1}}
\begin{tabular}{@{}R{2.5cm}lllR{1.25cm}llllR{1.82cm}@{}}
\toprule
& Known & Moves & Non-reg. & Target & 2-sided & Cited & Check & Mode & Sub. \\
\midrule
\multicolumn{10}{@{}l}{\textit{Benchmarks: A system is scored against mathematical problems}}\\
\addlinespace[1pt]
miniF2F, ProofNet, PutnamBench, FormalMATH, Lean Workbook & yes & no & NA & prop. & NA & NA & proof & auto & proof \\
CombiBench & yes & no & NA & prop. & NA & NA & proof & auto & answer + proof \\
SorryDB, ArXivLean & yes$^{c}$ & no & NA & prop. & NA & NA & proof & auto & proof \\
FrontierMath Erd\H{o}s & no & no & NA & prop. & NA & NA & proof & auto & proof \\
Formal Conjectures & no & no & no & function$^{a}$ & yes & yes$^{d}$ & proof & auto & answer term$^{a}$ \\
HorizonMath & no & no$^{b}$ & NA & scalar & no & yes & exec. & auto & Python program \\
FM: Open Problems & no & no & NA & scalar & no & yes & exec. & auto & real value \\ \hline
\addlinespace[1pt]
\multicolumn{10}{@{}l}{\textit{Records: An archive of the community's best known results}}\\
\addlinespace[1pt]
Ramsey, Cages & no & yes & yes & table & yes & yes & none & human & citation \\
LoB, Covering, Grassl & no & yes & yes & table & yes & yes & exec. & human & construction \\
Opt.\ Constants & no & yes & yes & scalar & yes & yes & none & human & a claim + link \\
EinsteinArena & no & yes & yes & scalar & no & yes & exec. & auto & a witness \\
FrontierShift & no & yes & yes & scalar$^{e}$ & no & yes & exec. & human & a witness \\
\hline
\addlinespace[1pt]
\textbf{SqueezeBench} & no & yes & yes & \textbf{function} & \textbf{yes} & \textbf{no} & proof & auto & \textbf{function + proofs} \\
\bottomrule
\end{tabular}

\end{table}

Benchmarks of solved problems, such as miniF2F~\citep{zheng2022minif2f} and ProofNet~\citep{azerbayev2023proofnet}, hold out known answers, which can leak into training data, and they saturate. Benchmarks of open problems avoid known answers. FrontierMath Erd\H{o}s~\citep{adamczewski2026frontiermatherdos} asks for Lean proofs or disproofs of 68 Erd\H{o}s problems, and Formal Conjectures~\citep{firsching2026formalconjectures} collects open conjectures from many areas, such as number theory and spectral theory, and HorizonMath~\citep{wang2026horizonmath} scores programs that compute values for open problems against fixed baselines.

Archives of best known results track moving targets, such as the surveys of small Ramsey numbers~\citep{radziszowski2026ramsey} and of cages~\citep{exoo2013cages}. Their updates rest on human checking, and each covers its own problems. Newer collections gather many problems in one place, such as Leaps in Bounds (LoB)~\citep{leapsinbounds2026}, FrontierShift~\citep{aitonomy2026openproblems} and Optimization Constants in Mathematics~\citep{davis2026optimizationconstants}, which Terence Tao started\footnote{https://terrytao.wordpress.com/2026/01/22/a-crowdsourced-repository-for-optimization-constants/}. They are updated semi-automatically and accept a claim backed by a citation, without a formal proof. FunSearch~\citep{romeraparedes2024funsearch}, AlphaEvolve~\citep{novikov2025alphaevolve} and PatternBoost~\citep{charton2024patternboost} find constructions by search and check each with a program; FunSearch's cap sets of growth $2.2202$ exceed our lower record. These systems are potential submitters, not competitors, since a construction enters the benchmark once proved in Lean. \SB is not a search method: it adds a Lean proof valid at every $n$ and keeps a lower and an upper record.

Progress on such problems rarely comes as a single proof. In combinatorics in particular, the bounds on an extremal function are tightened step by step by many researchers, and even an improvement at a single value can matter. A benchmark that asks for a proof of a fixed statement or for a better scalar cannot credit such steps on a function whose values are unknown, and a single proof of a conjecture does not credit the earlier work it builds on.

The columns of \Cref{tab:related} record whether the answers are \emph{Known}, whether a better submission replaces the incumbent (\emph{Moves}), whether the incumbent can never get worse (\emph{Non-reg.}), the \emph{Target} (a proposition, a scalar or a function of $n$), whether a lower and an upper bound are both kept (\emph{2-sided}), whether trust rests on a citation (\emph{Cited}), what performs the \emph{Check} (a kernel, or a program run on a witness), whether the check is automatic or human (\emph{Mode}), and the submission (\emph{Sub.}).

\section{Problems}
\label{sec:problems}

We include ten problems in \SB, but the list is not closed. We will keep adding problems that meet the same requirements as the current ones (\Cref{sec:framework}), and we welcome proposals of new problems from the community. Appendix~\ref{sec:adding} describes how a new problem could be added.

\subsection{The ten problems}
\label{sec:problem-list}

The ten problems come from three areas of extremal combinatorics (\Cref{tab:problems}): additive combinatorics (cap sets, Roth numbers, the Ruzsa--Szemer\'edi problem and Golomb rulers), extremal graph theory (graphs without a 4-cycle, graphs without cycles of length~3 or 4, and subgraphs of the hypercube without a 4-cycle), and Ramsey theory (the diagonal Ramsey numbers $R(n,n)$, the Ramsey numbers $R(3,k)$ and the Schur numbers). Each problem has a single integer parameter, and its exact values are known only up to a small size, for example $n\le 4$ for $R(n,n)$, $k\le 5$ for the Schur numbers, $n\le 6$ for cap sets and for the hypercube, and $n\le 28$ for Golomb rulers.

\begin{table}[!h]
\centering
\caption{The ten problems and their standing records after the runs of
\Cref{sec:evaluation}. Seed records are omitted; each is the simplest provable bound, and several are trivial.
Golomb is a minimization problem, and its upper record and lower record are the construction and
impossibility argument, respectively. For all other problems, constructions give the lower record.
$^{\star}$~matches the leading term of the best published bound, or its base if the bound is exponential, in that
direction. $^{\dagger}$~matches the best elementary bound; the published frontier uses the
triangle removal lemma. Unmarked records are asymptotically behind the literature.}
\label{tab:problems}
\footnotesize
\setlength{\tabcolsep}{2pt}
\renewcommand{\arraystretch}{1.12}
\hyphenpenalty=10000 \exhyphenpenalty=10000
\newcolumntype{Q}[1]{>{\raggedright\arraybackslash}p{#1}}
\begin{tabular}{@{}Q{2.0cm}Q{3.7cm}lQ{3.25cm}Q{3.25cm}@{}}
\toprule
Problem & $f$ counts & Domain & Standing lower record & Standing upper record \\
\midrule
\multicolumn{5}{@{}l}{\textit{Additive combinatorics}}\\
\addlinespace[1pt]
Cap set & largest subset of $\mathbb{F}_3^n$ with no three collinear points & $n\ge 1$
  & growth $112^{1/6}\approx 2.196$ (extended products of caps) & $O(2.756^n)$ (Ellenberg--Gijswijt)\,\textsuperscript{$\star$} \\
Roth & largest subset of $\{0,\dots,n-1\}$ with no 3-term progression & $n\ge 1$
  & $\Theta(n^{0.638})$; exact for $n\le 209$ & $4n/11+O(1)$ (Moser) \\
Ruzsa--\allowbreak Szemer\'edi & most triangles in an~$n$-vertex graph whose every edge lies in exactly one triangle & $n\ge 3$
  & $\Theta(n^{1.716})$ (from cap sets) & $n^2/18+O(n)$, smaller for $n\ge 33$ (counting)\,\textsuperscript{$\dagger$} \\
Golomb & shortest ruler with~$n$ marks and all differences distinct & $n\ge 1$
  & $n^2-2n^{3/2}+O(n)$\,\textsuperscript{$\star$} & $\approx n^2$ (Singer)\,\textsuperscript{$\star$}; best known for $n\le 100$ \\ \hline
\addlinespace[2pt]
\multicolumn{5}{@{}l}{\textit{Extremal graph theory}}\\
\addlinespace[1pt]
$C_4$-free & most edges in an~$n$-vertex graph with no 4-cycle & $n\ge 1$
  & $\approx\tfrac12 n^{3/2}$ (sub-polarity graphs)\,\textsuperscript{$\star$} & Reiman's bound $-\,O(\sqrt n)$\,\textsuperscript{$\star$} \\
Girth five & most edges in an~$n$-vertex graph with no 3- or 4-cycle & $n\ge 1$
  & $\Theta(n^{3/2})$ for all~$n$ (sub-planes of projective planes) & Moore bound $+\,O(1)$\,\textsuperscript{$\star$} \\
Hypercube $C_4$ & most edges in a $C_4$-free subgraph of the~$n$-cube & $n\ge 1$
  & $(n+\Theta(\sqrt n))\,2^{n-2}$ (Brass--\allowbreak Harborth--\allowbreak Nienborg)\,\textsuperscript{$\star$} & $\approx 0.623$ of the edges (Chung) \\ \hline
\addlinespace[2pt]
\multicolumn{5}{@{}l}{\textit{Ramsey theory}}\\
\addlinespace[1pt]
Ramsey & least~$N$ such that every 2-coloring of $K_N$ has a monochromatic $K_n$ & $n\ge 2$
  & $(1+o(1))\frac{\sqrt2}{e}\,n\,2^{n/2}$ (Spencer)\,\textsuperscript{$\star$}; best known at $n=5,6$ & Erd\H{o}s--Szekeres with Walker's step \\
$R(3,k)$ & least~$N$ such that every~$N$-vertex graph has a triangle or~$k$ independent vertices & $k\ge 2$
  & $\approx (k/2)^{3/2}$ (ordered point--line incidences) & $(1+o(1))\,k^2/\log k$ (Shearer)\,\textsuperscript{$\star$} \\
Schur & largest~$N$ such that $\{1,\dots,N\}$ has a~$k$-coloring with no monochromatic $x+y=z$ & $k\ge 1$
  & growth $380^{1/5}\approx 3.28$ (template constructions)\,\textsuperscript{$\star$} & $\lfloor e\,k!\rfloor-k$ for odd~$k$ \\
\bottomrule
\end{tabular}
\end{table}

\subsection{Running example: Golomb rulers}
\label{sec:golomb}

The values of $f(n)$ for the Golomb ruler problem of \Cref{sec:framework} are known for $n\le 28$; the last five were settled by exhaustive searches by distributed.net~\citep{oeisA003022}. The lower seed is $\binom{n}{2}$, since its $\binom{n}{2}$ distances are distinct and at most the length. The upper seed is the shorter of the binary ruler $\{0,1,3,\dots,2^{n-1}-1\}$ and a ruler of length about $2n^3$, patched at $n=5,6$.

\begin{figure}[t]
\centering
\begin{minipage}{0.97\textwidth}
\footnotesize
\begin{alltt}
\textbf{def} IsSidon (S : Finset \(\mathbb{N}\)) : Prop :=
  \(\forall\) a \(\in\) S, \(\forall\) b \(\in\) S, \(\forall\) c \(\in\) S, \(\forall\) d \(\in\) S,
    a + b = c + d \(\to\) (a = c \(\land\) b = d) \(\lor\) (a = d \(\land\) b = c)
\textbf{def} RulerLengths (m : \(\mathbb{N}\)) : Set \(\mathbb{N}\) :=
  \{L | \(\exists\) S : Finset \(\mathbb{N}\), S \(\subseteq\) range (L + 1) \(\land\) S.card = m \(\land\)
        IsSidon S \(\land\) L \(\in\) S\}
\textbf{noncomputable def} f (m : \(\mathbb{N}\)) : \(\mathbb{N}\) :=
  if m < nMin then 0 else sInf (RulerLengths m)
\end{alltt}
\end{minipage}
\caption{The Lean definition of~$f$ for Golomb rulers (parameter~$m$, $n_{\min}=1$).}
\label{fig:golomb-lean}
\end{figure}

\Cref{fig:golomb-lean} shows how~$f$ is formalized. Like every problem, it builds on Mathlib~\citep{mathlib2020}, and four problems reuse a Mathlib definition of~$f$ directly, such as \texttt{rothNumberNat} for Roth numbers. The ruler condition is stated through sums: in a Sidon set, $a+b=c+d$ determines the pair, which for a finite set of integers is equivalent to all positive differences being distinct. Since~$f$ is an infimum and the infimum of the empty set is 0 in Lean, the file proves that the binary ruler above is a ruler; without it, no lower bound on~$f$ could be proved. The definition is \texttt{noncomputable}: nothing in the pipeline evaluates~$f$, and a record may not refer to~$f$ or \texttt{RulerLengths}, since it would compute~$f$ under another name.

\section{Evaluation}
\label{sec:evaluation}

On the ten problems included as part of \SB, we performed an agent-based evaluation that provides a baseline and a continuation from its records. It has two phases: 

\begin{enumerate}
  \item We obtained preliminary bounds using Claude Code agents with the model Opus 5, starting from elementary seeds under a budget of one hour (or until two idle sessions in a row) and 16 GiB of memory. They ran in parallel on an AMD Ryzen 9 5900X machine with 24 threads at 3.7 GHz and 125 GiB of memory under Ubuntu 22.04.5 LTS.
  \item From the bounds of the first phase, we launched Fable 5.1 agents at maximum effort on 19 of the 20 records, and an Opus 5.5 agent on the lower record for Golomb rulers, with the possibility of consulting GPT-6 Astra through its API. Each agent had at most 32 GiB of memory, and each run took up to about two hours. We sent a stop notice after 80 minutes or at 55\% of the context, and a notice to continue when an agent stopped earlier; the first six runs ended before this rule existed. The prompt changed slightly over the runs, and a rule on outside sources was added during them. The agents ran on two AMD EPYC 9274F CPUs (96 threads) with 755 GiB of memory under Ubuntu 24.04.
\end{enumerate}

In both phases, one agent was used per track (10 problems $\times$ 2 directions). They were all prompted to first read the documentation about the problem and the current records, possibly search the literature and choose a bound. Then they had to develop a Lean proof of the \texttt{valid}, \texttt{improves} and \texttt{strict} theorems, run \texttt{squeeze verify} and, if it passed, accept the bound as a new record and repeat the loop. The agents accepted records in their own copies of the repository. We re-verified every record of the second phase independently before merging it, and later re-ran \texttt{squeeze audit} on every record of both phases. The prompt also asks for improvements at infinitely many~$n$ and discourages many tiny submissions. The repository contains the full prompt.

Every one of the twenty records improved on its seed. In the first phase, the Opus~5 agents had 82 submissions accepted in 18.9 hours, and sixteen of the twenty tracks were stopped by the time limit rather than by two idle sessions. In the second phase, the Fable~5.1 and Opus~5.5 agents had 60 more submissions accepted and improved 19 of the 20 records again. \Cref{tab:problems} lists the standing records: ten match the leading term or the growth rate of the best published bound, one matches the best elementary bound, and nine are asymptotically behind. Nearly all accepted proofs are general arguments, such as explicit algebraic constructions, products of smaller constructions, and counting arguments; \texttt{decide} only checks explicit witnesses. Gate 6 was added during the second phase, and it rejected one record that had passed the earlier checks in an agent's copy of the repository but was not yet merged. The first formalization of the Ellenberg--Gijswijt bound computed its values by a recursion over lists and exceeded the budget at every~$n$ from 59 to 100, so the next submitter could not have evaluated it. The agent resubmitted the bound with a computation that needs fewer than 100 heartbeats per~$n$. One record may go beyond the literature at concrete values. For the Ruzsa--Szemer\'edi problem, a Fable~5.1 agent localized a counting argument of \citet{fox2017induced}, suggested by the second model it consulted, and combined it with the degree bounds of earlier versions. The resulting upper record gives $f(50)\le 140$ and $f(100)\le 529$, where the best bounds we could derive from the literature are 144 and 566. We found no published bound that gives these values; asymptotically, the record is still far from the $o(n^2)$ bound of the triangle removal lemma.

The running example shows the difference between the two phases (\Cref{fig:gap}). Opus~5 reached the Erd\H{o}s--Tur\'an ruler, of length about $2n^2$, with the optimal rulers for $n\le 17$, and refined counting arguments over the distances of a ruler to $n^2-2n^{3/2}+O(n)$, the form of Lindstr\"om's bound~\citep{lindstrom1969}. Fable~5.1 brought in constructions from the literature, down to length about $n^2$~\citep{singer1938,bose1962,ruzsa1993}, and tables of explicit rulers, the optimal ones for $n\le 28$ and the best known ones up to $n=100$~\citep{rokicki2016golomb}; the kernel checks each ruler. The upper record went from 1570 to the optimum 585 at $n=28$, and from 20{,}002 to 8831 at $n=100$. Three submissions by Opus~5.5 refined the lower-order terms of the lower record, from 529 to 536 at $n=28$ and from 8143 to 8210 at $n=100$. In total, 13 upper and 8 lower versions were accepted. The records now determine $f(n)$ for $n\le 5$, and the gap is 49 at $n=28$ and 621 at $n=100$, where the seeds left 42{,}687 and 1{,}984{,}851. Both records match the leading term $n^2$ of the best published bounds and go no further; the published lower bound has a better second-order term~\citep{balogh2023sidon}.

Apart from possibly the Ruzsa--Szemer\'edi record, no record goes beyond the published literature. Both records match the leading term of the best published bounds only for $C_4$-free graphs and Golomb rulers; for the other eight problems at least one record stays behind (\Cref{tab:problems}). For example, the upper record for the diagonal Ramsey numbers gives $R(5,5)\le 58$, while the published bound is 46~\citep{radziszowski2026ramsey}. Some of the best bounds rest on results that Mathlib lacks, or whose explicit constants are far too weak at these~$n$, such as the triangle removal lemma. Others were obtained by computer, and there the two directions differ. A construction found by a search comes with a witness, which is cheap to check once it is published: the agents imported the best known Golomb rulers up to 100 marks and a Schur coloring that shows $S(7)\ge 1696$. An impossibility result found by an exhaustive search has no short certificate and exceeds the computation budget, as for $f(28)=585$, so it can enter a record only through an argument that is not known yet. Published but unformalized bounds leave room for further submissions. Beyond them, the records reach values of~$f$ that nobody knows, and formalizing the literature alone cannot saturate the benchmark.

\section{Conclusions}

\SB measures mathematical progress by records rather than by solved statements. A record is a computable function with a proof that it bounds~$f$, and a submission replaces it only by proving that it is valid, no worse at any~$n$ and better at some~$n$, checked by two kernels against statements fixed in advance. In our evaluation the records left by one system became the starting point of the next. After 82 accepted submissions of the baseline, agents based on newer models had 60 more accepted and improved 19 of the 20 records again, and eight of the ten problems still have a record behind the literature. The record history is the leaderboard.

The benchmark has limits. Its ten problems come from one area of mathematics and have one parameter each. Our runs were not a controlled comparison of models from one snapshot with equal budgets, which is how we intend the benchmark to be used. A maintainer still reads each report for a search above $n=100$. Bounds known only from exhaustive computer searches cannot enter a record within the budget. Future work will add problems with several parameters and admit proofs that rely on a verified checker for SAT certificates, marked as such, as a route for these bounds. %
\SB is open to submissions from people and AI systems alike.

\needspace{10\baselineskip}
\section*{Acknowledgements}

This research was funded in whole or in part by the Austrian Science Fund (FWF) within the Cluster of Excellence Bilateral Artificial Intelligence (10.55776/COE12).

\medskip
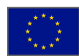
\begin{wrapfigure}[2]{l}{25pt}\vspace{-8pt}{\worldflag[width=18pt]{EU}}
\end{wrapfigure}
\noindent This project is also partially supported by the European Union's Horizon 2020 research and innovation programme
under the Marie Sk\l{}odowska-Curie grant agreement No.~101034440.

\section*{Use of AI assistants}
We used generative AI tools to implement the benchmark tools and the Lean formalizations, to give feedback on the design of the admission checks, and to compare the records with the literature. The records of \Cref{sec:evaluation} were produced by the AI agents under evaluation. The other tasks with required disclosure do not apply to this work. We also used AI tools to draft and edit text and figures, search the literature and format references. We reviewed all AI-assisted work. Code and proofs are checked by tests and the Lean kernel, and every cited source was checked against its original. We take responsibility for the final content of this paper, including text, claims and artifacts produced with the aid of generative AI.

\bibliography{references}

\begin{thebibliography}{45}
\providecommand{\natexlab}[1]{#1}
\providecommand{\url}[1]{\texttt{#1}}
\expandafter\ifx\csname urlstyle\endcsname\relax
  \providecommand{\doi}[1]{doi: #1}\else
  \providecommand{\doi}{doi: \begingroup \urlstyle{rm}\Url}\fi

\bibitem[Acerbi(2026)]{coveringrepository2026}
G.~Acerbi.
\newblock {Covering Repository}.
\newblock \url{https://www.coveringrepository.com/}, 2026.
\newblock Incorporates the data of D.~Gordon's {La Jolla} Covering Repository
  up to March 2026; updates dating to November 1996; distributes validation
  software. Accessed 2026-09-21.

\bibitem[Adamczewski and Bloom(2026)]{adamczewski2026frontiermatherdos}
T.~Adamczewski and T.~F. Bloom.
\newblock {FrontierMath} {Erd\H{o}s}, 2026.
\newblock URL \url{https://arxiv.org/abs/2609.25050}.
\newblock 68 {Erd\H{o}s} problems open as of August 2026; resolution required
  in {Lean}.

\bibitem[{AItonomy Foundation}(2026)]{aitonomy2026openproblems}
{AItonomy Foundation}.
\newblock {FrontierShift}: Open problems.
\newblock \url{https://aitonomy.org/projects/open-problems}, 2026.
\newblock 167 problems; incumbent-beating submission rule; per-problem
  lower/upper bounds and gap. Accessed 2026-09-21.

\bibitem[Akhtar et~al.(2026)Akhtar, Reuel, Soni, Ahuja, Ammanamanchi, Rawal,
  Zouhar, Yadav, Whitehouse, Ki, Mickel, Choshen, Suppa, Batzner, Chim, Sania,
  Long, Rahmani, Knight, Nan, Raj, Fan, Singh, Sahoo, Habba, Gohar, Pawar,
  Scholz, Subramonian, Ni, Kochenderfer, Koyejo, Sachan, Biderman, Talat,
  Ghosh, and Solaiman]{akhtar2026plateau}
M.~Akhtar, A.~Reuel, P.~Soni, S.~Ahuja, P.~S. Ammanamanchi, R.~Rawal,
  V.~Zouhar, S.~Yadav, C.~Whitehouse, D.~Ki, J.~Mickel, L.~Choshen, M.~Suppa,
  J.~Batzner, J.~Chim, J.~Sania, Y.~Long, H.~A. Rahmani, C.~Q. Knight, Y.~Nan,
  J.~Raj, Y.~Fan, S.~Singh, S.~Sahoo, E.~Habba, U.~Gohar, S.~M. Pawar,
  R.~Scholz, A.~Subramonian, J.~Ni, M.~Kochenderfer, S.~Koyejo, M.~Sachan,
  S.~Biderman, Z.~Talat, A.~Ghosh, and I.~Solaiman.
\newblock When {AI} benchmarks plateau: A systematic study of benchmark
  saturation.
\newblock In \emph{Forty-third International Conference on Machine Learning},
  2026.
\newblock URL \url{https://openreview.net/forum?id=YC1Otscjbs}.

\bibitem[Azerbayev et~al.(2023)Azerbayev, Piotrowski, Schoelkopf, Ayers, Radev,
  and Avigad]{azerbayev2023proofnet}
Z.~Azerbayev, B.~Piotrowski, H.~Schoelkopf, E.~W. Ayers, D.~Radev, and
  J.~Avigad.
\newblock {ProofNet}: Autoformalizing and formally proving undergraduate-level
  mathematics.
\newblock \emph{CoRR}, abs/2302.12433, 2023.
\newblock \doi{10.48550/ARXIV.2302.12433}.
\newblock URL \url{https://doi.org/10.48550/arXiv.2302.12433}.

\bibitem[Bailey(2026)]{nanoda}
C.~Bailey.
\newblock nanoda\_lib: a type checker for {Lean}~4.
\newblock \url{https://github.com/ammkrn/nanoda_lib}, 2026.
\newblock Accessed 2026-09-25.

\bibitem[Balogh et~al.(2023)Balogh, F{\"{u}}redi, and Roy]{balogh2023sidon}
J.~Balogh, Z.~F{\"{u}}redi, and S.~Roy.
\newblock An upper bound on the size of {Sidon} sets.
\newblock \emph{Am. Math. Mon.}, 130\penalty0 (5):\penalty0 437--445, 2023.
\newblock \doi{10.1080/00029890.2023.2176667}.
\newblock URL \url{https://doi.org/10.1080/00029890.2023.2176667}.

\bibitem[Bianchi et~al.(2026)Bianchi, Kwon, Pappu, and
  Zou]{bianchi2026einsteinarena}
F.~Bianchi, Y.~Kwon, A.~Pappu, and J.~Zou.
\newblock Harnessing the collective intelligence of {AI} agents in the wild for
  new discoveries.
\newblock \emph{CoRR}, abs/2606.10402, 2026.
\newblock \doi{10.48550/ARXIV.2606.10402}.
\newblock URL \url{https://doi.org/10.48550/arXiv.2606.10402}.

\bibitem[Bloom and Golomb(1977)]{bloom1977golomb}
G.~S. Bloom and S.~W. Golomb.
\newblock Applications of numbered undirected graphs.
\newblock \emph{Proceedings of the IEEE}, 65\penalty0 (4):\penalty0 562--570,
  1977.
\newblock \doi{10.1109/PROC.1977.10517}.

\bibitem[Bose and Chowla(1962)]{bose1962}
R.~C. Bose and S.~Chowla.
\newblock Theorems in the additive theory of numbers.
\newblock \emph{Commentarii Mathematici Helvetici}, 37\penalty0 (1):\penalty0
  141--147, 1962.
\newblock \doi{10.1007/BF02566968}.

\bibitem[Charton et~al.(2024)Charton, Ellenberg, Wagner, and
  Williamson]{charton2024patternboost}
F.~Charton, J.~S. Ellenberg, A.~Z. Wagner, and G.~Williamson.
\newblock {PatternBoost}: Constructions in mathematics with a little help from
  {AI}.
\newblock \emph{CoRR}, abs/2411.00566, 2024.
\newblock \doi{10.48550/ARXIV.2411.00566}.
\newblock URL \url{https://doi.org/10.48550/arXiv.2411.00566}.

\bibitem[Chen et~al.(2025)Chen, Gu, Huang, Huang, Jiang, Jie, Jin, Jin, Li, Ma,
  Ren, Shen, Shi, Sun, Sun, Wang, Wang, Wang, Wei, Wei, Wu, Wu, Xia, Xin, Yang,
  Ying, Yuan, Yuan, Zhan, Zhang, Zhang, Zhang, Zhao, Zhao, Zhou, and
  Zhu]{chen2025seedprover}
L.~Chen, J.~Gu, L.~Huang, W.~Huang, Z.~Jiang, A.~Jie, X.~Jin, X.~Jin, C.~Li,
  K.~Ma, C.~Ren, J.~Shen, W.~Shi, T.~Sun, H.~Sun, J.~Wang, S.~Wang, Z.~Wang,
  C.~Wei, S.~Wei, Y.~Wu, Y.~Wu, Y.~Xia, H.~Xin, F.~Yang, H.~Ying, H.~Yuan,
  Z.~Yuan, T.~Zhan, C.~Zhang, Y.~Zhang, G.~Zhang, T.~Zhao, J.~Zhao, Y.~Zhou,
  and T.~H. Zhu.
\newblock {Seed-Prover}: Deep and broad reasoning for automated theorem
  proving.
\newblock \emph{CoRR}, abs/2507.23726, 2025.
\newblock \doi{10.48550/ARXIV.2507.23726}.
\newblock URL \url{https://doi.org/10.48550/arXiv.2507.23726}.

\bibitem[Davis et~al.(2026)Davis, Ivanisvili, and
  Tao]{davis2026optimizationconstants}
D.~Davis, P.~Ivanisvili, and T.~Tao.
\newblock {Optimization Constants in Mathematics}.
\newblock \url{https://teorth.github.io/optimizationproblems/}, 2026.
\newblock Announced 22 January 2026. Accessed 2026-09-21.

\bibitem[de~Moura and Ullrich(2021)]{moura2021lean4}
L.~de~Moura and S.~Ullrich.
\newblock The {Lean} 4 theorem prover and programming language.
\newblock In A.~Platzer and G.~Sutcliffe, editors, \emph{Automated Deduction -
  {CADE} 28 - 28th International Conference on Automated Deduction, Virtual
  Event, July 12-15, 2021, Proceedings}, volume 12699 of \emph{Lecture Notes in
  Computer Science}, pages 625--635. Springer, 2021.
\newblock \doi{10.1007/978-3-030-79876-5\_37}.
\newblock URL \url{https://doi.org/10.1007/978-3-030-79876-5\_37}.

\bibitem[{Epoch AI}(2026)]{epoch2026frontiermathopen}
{Epoch AI}.
\newblock {FrontierMath}: Open problems.
\newblock \url{https://epoch.ai/frontiermath/open-problems}, 2026.
\newblock Pilot January 2026; expanded to 50 problems 31 July 2026. Accessed
  2026-09-21.

\bibitem[Erd{\H{o}}s and Tur{\'a}n(1941)]{erdos1941sidon}
P.~Erd{\H{o}}s and P.~Tur{\'a}n.
\newblock On a problem of {Sidon} in additive number theory, and on some
  related problems.
\newblock \emph{Journal of the London Mathematical Society}, s1-16\penalty0
  (4):\penalty0 212--215, 1941.
\newblock \doi{10.1112/jlms/s1-16.4.212}.

\bibitem[Exoo and Jajcay(2013)]{exoo2013cages}
G.~Exoo and R.~Jajcay.
\newblock Dynamic cage survey.
\newblock \emph{The Electronic Journal of Combinatorics}, 2013.
\newblock \doi{10.37236/37}.
\newblock Dynamic Survey DS16; first published 2008, revision 26 July 2013.

\bibitem[Feng et~al.(2026)Feng, Trinh, Bingham, Kang, Zhang, Kim, Barreto,
  Schildkraut, Jung, Seo, Pagano, Chervonyi, Hwang, Hou, Gukov, Tsai, Choi,
  Jin, Li, Wu, Shiu, Shih, Le, and Luong]{feng2026erdos}
T.~Feng, T.~H. Trinh, G.~Bingham, J.~Kang, S.~Zhang, S.~Kim, K.~Barreto,
  C.~Schildkraut, J.~Jung, J.~Seo, C.~Pagano, Y.~Chervonyi, D.~Hwang, K.~Hou,
  S.~Gukov, C.~Tsai, H.~Choi, Y.~Jin, W.~Li, H.~Wu, R.~Shiu, Y.~Shih, Q.~V. Le,
  and T.~Luong.
\newblock Semi-autonomous mathematics discovery with {Gemini}: {A} case study
  on the {Erd{\H{o}}s} problems.
\newblock \emph{CoRR}, abs/2601.22401, 2026.
\newblock \doi{10.48550/ARXIV.2601.22401}.
\newblock URL \url{https://doi.org/10.48550/arXiv.2601.22401}.

\bibitem[Firsching et~al.(2026)Firsching, Lezeau, Mercuri, Horv{\'{a}}th,
  Dillies, S{\"{o}}nne, Wieser, Zhang, Hubert, y~Arcas, and
  Kohli]{firsching2026formalconjectures}
M.~Firsching, P.~Lezeau, S.~Mercuri, M.~Z. Horv{\'{a}}th, Y.~Dillies,
  C.~S{\"{o}}nne, E.~Wieser, F.~Zhang, T.~Hubert, B.~A. y~Arcas, and P.~Kohli.
\newblock {Formal Conjectures}: An open and evolving benchmark for verified
  discovery in mathematics.
\newblock \emph{CoRR}, abs/2605.13171, 2026.
\newblock \doi{10.48550/ARXIV.2605.13171}.
\newblock URL \url{https://doi.org/10.48550/arXiv.2605.13171}.

\bibitem[Fox et~al.(2017)Fox, Huang, and Sudakov]{fox2017induced}
J.~Fox, H.~Huang, and B.~Sudakov.
\newblock On graphs decomposable into induced matchings of linear sizes.
\newblock \emph{Bulletin of the London Mathematical Society}, 49\penalty0
  (1):\penalty0 45--57, 2017.
\newblock \doi{10.1112/blms.12005}.

\bibitem[Gehrunger et~al.(2026)Gehrunger, Dekoninck, and
  Vechev]{matharena2026arxivlean}
T.~Gehrunger, J.~Dekoninck, and M.~Vechev.
\newblock {ArXivLean}: How well can {LLMs} formally prove research math?
\newblock \url{https://matharena.ai/arxivlean/}, 2026.
\newblock First public release 21 April 2026. Accessed 2026-09-21.

\bibitem[Grassl(2026)]{grassl2026codetables}
M.~Grassl.
\newblock Bounds on the minimum distance of linear codes and quantum codes.
\newblock \url{http://www.codetables.de}, 2026.
\newblock Accessed 2026-09-21.

\bibitem[{Lean FRO}(2026{\natexlab{a}})]{comparator}
{Lean FRO}.
\newblock Comparator.
\newblock \url{https://github.com/leanprover/comparator}, 2026{\natexlab{a}}.
\newblock Accessed 2026-09-25.

\bibitem[{Lean FRO}(2026{\natexlab{b}})]{leaneval}
{Lean FRO}.
\newblock lean-eval.
\newblock \url{https://github.com/leanprover/lean-eval}, 2026{\natexlab{b}}.
\newblock Accessed 2026-09-25.

\bibitem[Letson et~al.(2026)Letson, Sarra, Poiroux, Dressler, Lezeau, Aranha,
  Pu, Hill, Hidalgo, Berman, Tsoukalas, and Taelman]{letson2026sorrydb}
A.~Letson, L.~Sarra, A.~Poiroux, O.~Dressler, P.~Lezeau, D.~Aranha, F.~Pu,
  A.~Hill, M.~C. Hidalgo, J.~Berman, G.~Tsoukalas, and L.~Taelman.
\newblock {SorryDB}: Can {AI} provers complete real-world {Lean} theorems?
\newblock \emph{CoRR}, abs/2603.02668, 2026.
\newblock \doi{10.48550/ARXIV.2603.02668}.
\newblock URL \url{https://doi.org/10.48550/arXiv.2603.02668}.

\bibitem[Lindstr{\"o}m(1969)]{lindstrom1969}
B.~Lindstr{\"o}m.
\newblock An inequality for {$B_2$}-sequences.
\newblock \emph{Journal of Combinatorial Theory}, 6\penalty0 (2):\penalty0
  211--212, 1969.
\newblock \doi{10.1016/S0021-9800(69)80124-9}.

\bibitem[Liu et~al.(2025)Liu, Lin, Bayer, Dillies, Jiang, Liang, Soletskyi,
  Wang, Xie, Xiong, Yang, Zhang, Zhi, Li, and Liu]{liu2025combibench}
J.~Liu, X.~Lin, J.~Bayer, Y.~Dillies, W.~Jiang, X.~Liang, R.~Soletskyi,
  H.~Wang, Y.~Xie, B.~Xiong, Z.~Yang, J.~Zhang, L.~Zhi, J.~Li, and Z.~Liu.
\newblock {CombiBench}: Benchmarking {LLM} capability for combinatorial
  mathematics.
\newblock \emph{CoRR}, abs/2505.03171, 2025.
\newblock \doi{10.48550/ARXIV.2505.03171}.
\newblock URL \url{https://doi.org/10.48550/arXiv.2505.03171}.

\bibitem[Luong and Lockhart(2025)]{luong2025imo}
T.~Luong and E.~Lockhart.
\newblock Advanced version of {Gemini} with {Deep Think} officially achieves
  gold-medal standard at the {International Mathematical Olympiad}.
\newblock Google DeepMind blog, 2025.
\newblock URL
  \url{https://deepmind.google/blog/advanced-version-of-gemini-with-deep-think-officially-achieves-gold-medal-standard-at-the-international-mathematical-olympiad/}.
\newblock Accessed 2026-09-24.

\bibitem[Munro(2026)]{leapsinbounds2026}
D.~Munro.
\newblock {Leaps in Bounds}.
\newblock \url{https://leapsinbounds.org/}, 2026.
\newblock Per-parameter histories of lower and upper bounds. Accessed
  2026-09-21.

\bibitem[Novikov et~al.(2025)Novikov, Vu, Eisenberger, Dupont, Huang, Wagner,
  Shirobokov, Kozlovskii, Ruiz, Mehrabian, Kumar, See, Chaudhuri, Holland,
  Davies, Nowozin, Kohli, and Balog]{novikov2025alphaevolve}
A.~Novikov, N.~Vu, M.~Eisenberger, E.~Dupont, P.~Huang, A.~Z. Wagner,
  S.~Shirobokov, B.~Kozlovskii, F.~J.~R. Ruiz, A.~Mehrabian, M.~P. Kumar,
  A.~See, S.~Chaudhuri, G.~Holland, A.~Davies, S.~Nowozin, P.~Kohli, and
  M.~Balog.
\newblock {AlphaEvolve}: {A} coding agent for scientific and algorithmic
  discovery.
\newblock \emph{CoRR}, abs/2506.13131, 2025.
\newblock \doi{10.48550/ARXIV.2506.13131}.
\newblock URL \url{https://doi.org/10.48550/arXiv.2506.13131}.

\bibitem[{OEIS Foundation Inc.}(2026)]{oeisA003022}
{OEIS Foundation Inc.}
\newblock Sequence {A003022}: Length of shortest (or optimal) {Golomb} ruler
  with $n$ marks.
\newblock The On-Line Encyclopedia of Integer Sequences,
  \url{https://oeis.org/A003022}, 2026.
\newblock Accessed 2026-09-25.

\bibitem[{Palomar}(2026)]{palomar}
{Palomar}.
\newblock {Palomar}: A registry of {Lean}-verified mathematics.
\newblock \url{https://palomar-registry.org/about}, 2026.
\newblock Accessed 2026-09-25.

\bibitem[{PutnamBench maintainers}(2026)]{putnambench2026leaderboard}
{PutnamBench maintainers}.
\newblock {PutnamBench} leaderboard, 2026.
\newblock URL \url{https://trishullab.github.io/PutnamBench/leaderboard.html}.
\newblock Accessed 2026-09-24.

\bibitem[Radziszowski(2026)]{radziszowski2026ramsey}
S.~P. Radziszowski.
\newblock Small {Ramsey} numbers.
\newblock \emph{The Electronic Journal of Combinatorics}, 2026.
\newblock \doi{10.37236/21}.
\newblock Dynamic Survey DS1; first published 1994, revision \#18, 24 April
  2026.

\bibitem[Rokicki and Dogon(2016)]{rokicki2016golomb}
T.~Rokicki and G.~Dogon.
\newblock Possibly optimal {Golomb} rulers calculated for 160 to 40,000 marks.
\newblock \url{https://cube20.org/golomb/}, 2016.
\newblock Accessed 2026-09-25.

\bibitem[Romera{-}Paredes et~al.(2024)Romera{-}Paredes, Barekatain, Novikov,
  Balog, Kumar, Dupont, Ruiz, Ellenberg, Wang, Fawzi, Kohli, and
  Fawzi]{romeraparedes2024funsearch}
B.~Romera{-}Paredes, M.~Barekatain, A.~Novikov, M.~Balog, M.~P. Kumar,
  E.~Dupont, F.~J.~R. Ruiz, J.~S. Ellenberg, P.~Wang, O.~Fawzi, P.~Kohli, and
  A.~Fawzi.
\newblock Mathematical discoveries from program search with large language
  models.
\newblock \emph{Nat.}, 625\penalty0 (7995):\penalty0 468--475, 2024.
\newblock \doi{10.1038/S41586-023-06924-6}.
\newblock URL \url{https://doi.org/10.1038/s41586-023-06924-6}.

\bibitem[Ruzsa(1993)]{ruzsa1993}
I.~Z. Ruzsa.
\newblock Solving a linear equation in a set of integers {I}.
\newblock \emph{Acta Arithmetica}, 65\penalty0 (3):\penalty0 259--282, 1993.
\newblock \doi{10.4064/aa-65-3-259-282}.

\bibitem[Singer(1938)]{singer1938}
J.~Singer.
\newblock A theorem in finite projective geometry and some applications to
  number theory.
\newblock \emph{Transactions of the American Mathematical Society}, 43\penalty0
  (3):\penalty0 377--385, 1938.
\newblock \doi{10.1090/S0002-9947-1938-1501951-4}.

\bibitem[Stewart(2026)]{epoch2026brief}
E.~Stewart.
\newblock The {Epoch Brief}, {September} 12, 2026.
\newblock Epoch AI newsletter, 2026.
\newblock URL
  \url{https://epochai.substack.com/p/the-epoch-brief-september-12-2026}.
\newblock Accessed 2026-09-24.

\bibitem[{The mathlib Community}(2020)]{mathlib2020}
{The mathlib Community}.
\newblock The {Lean} mathematical library.
\newblock In J.~Blanchette and C.~Hritcu, editors, \emph{Proceedings of the 9th
  {ACM} {SIGPLAN} International Conference on Certified Programs and Proofs,
  {CPP} 2020, New Orleans, LA, USA, January 20-21, 2020}, pages 367--381.
  {ACM}, 2020.
\newblock \doi{10.1145/3372885.3373824}.
\newblock URL \url{https://doi.org/10.1145/3372885.3373824}.

\bibitem[Tsoukalas et~al.(2024)Tsoukalas, Lee, Jennings, Xin, Ding, Jennings,
  Thakur, and Chaudhuri]{tsoukalas2024putnambench}
G.~Tsoukalas, J.~Lee, J.~Jennings, J.~Xin, M.~Ding, M.~Jennings, A.~Thakur, and
  S.~Chaudhuri.
\newblock {PutnamBench}: Evaluating neural theorem-provers on the {Putnam
  Mathematical Competition}.
\newblock In A.~Globersons, L.~Mackey, D.~Belgrave, A.~Fan, U.~Paquet, J.~M.
  Tomczak, and C.~Zhang, editors, \emph{Advances in Neural Information
  Processing Systems 37: Annual Conference on Neural Information Processing
  Systems 2024, NeurIPS 2024, Vancouver, BC, Canada, December 10 - 15, 2024},
  2024.
\newblock URL
  \url{http://papers.nips.cc/paper\_files/paper/2024/hash/1582eaf9e0cf349e1e5a6ee453100aa1-Abstract-Datasets\_and\_Benchmarks\_Track.html}.

\bibitem[Wang et~al.(2026)Wang, Motwani, Roggeveen, Hodges, Jayalath, London,
  Ramakrishnan, Cipcigan, Torr, and Abate]{wang2026horizonmath}
E.~Y. Wang, S.~R. Motwani, J.~V. Roggeveen, E.~Hodges, D.~Jayalath, C.~London,
  K.~Ramakrishnan, F.~Cipcigan, P.~Torr, and A.~Abate.
\newblock {HorizonMath}: Measuring {AI} progress toward mathematical discovery
  with automatic verification.
\newblock \emph{CoRR}, abs/2603.15617, 2026.
\newblock \doi{10.48550/ARXIV.2603.15617}.
\newblock URL \url{https://doi.org/10.48550/arXiv.2603.15617}.

\bibitem[Ying et~al.(2024)Ying, Wu, Geng, Wang, Lin, and
  Chen]{ying2024leanworkbook}
H.~Ying, Z.~Wu, Y.~Geng, J.~Wang, D.~Lin, and K.~Chen.
\newblock {Lean Workbook}: {A} large-scale {Lean} problem set formalized from
  natural language math problems.
\newblock In A.~Globersons, L.~Mackey, D.~Belgrave, A.~Fan, U.~Paquet, J.~M.
  Tomczak, and C.~Zhang, editors, \emph{Advances in Neural Information
  Processing Systems 37: Annual Conference on Neural Information Processing
  Systems 2024, NeurIPS 2024, Vancouver, BC, Canada, December 10 - 15, 2024},
  2024.
\newblock URL
  \url{http://papers.nips.cc/paper\_files/paper/2024/hash/bf236666a2cc5f3ae05d2e08485efc4c-Abstract-Datasets\_and\_Benchmarks\_Track.html}.

\bibitem[Yu et~al.(2025)Yu, Peng, Ding, Li, Peng, Liu, Zhang, Yuan, Xin, Huang,
  Wen, Zhang, and Liu]{formalmath2025}
Z.~Yu, R.~Peng, K.~Ding, Y.~Li, Z.~Peng, M.~Liu, Y.~Zhang, Z.~Yuan, H.~Xin,
  W.~Huang, Y.~Wen, G.~Zhang, and W.~Liu.
\newblock {FormalMATH}: Benchmarking formal mathematical reasoning of large
  language models.
\newblock \emph{CoRR}, abs/2505.02735, 2025.
\newblock \doi{10.48550/ARXIV.2505.02735}.
\newblock URL \url{https://doi.org/10.48550/arXiv.2505.02735}.

\bibitem[Zheng et~al.(2022)Zheng, Han, and Polu]{zheng2022minif2f}
K.~Zheng, J.~M. Han, and S.~Polu.
\newblock {miniF2F}: a cross-system benchmark for formal {Olympiad}-level
  mathematics.
\newblock In \emph{The Tenth International Conference on Learning
  Representations, {ICLR} 2022, Virtual Event, April 25-29, 2022}.
  OpenReview.net, 2022.
\newblock URL \url{https://openreview.net/forum?id=9ZPegFuFTFv}.

\end{thebibliography}
\bibliographystyle{abbrvnat}

\appendix

\section{Adding a problem}
\label{sec:adding}

To add a new problem to \SB, the following steps must be followed.
\begin{enumerate}
\item \emph{Define the problem.} \texttt{Problems/<P>/Def.lean} defines $n_{\min}$ and $f$ by its combinatorial meaning, pinned to $0$ below $n_{\min}$, reusing a definition from Mathlib where one exists. It also provides, for each direction, a lemma that reduces a bound to its combinatorial content, such as the existence of a witness of a given size. Where $f$ can be evaluated, its smallest values are checked by \texttt{decide} on every build. If $f$ is a least element, the file also proves that the set is nonempty; without this, no lower record can be proved.
\item \emph{Seed both records.} Each direction gets a version V000 in \texttt{Records/<Dir>/V000/}, with the same four files as a submission: the record in \texttt{Def.lean}, a proof of \texttt{valid} alone in \texttt{Proof.lean}, \texttt{metadata.toml} and \texttt{REPORT.md}. A seed should be simple and quick to formalize, and it may not be taken from the known values.
\item \emph{Configure the checks.} A trusted file \texttt{Challenges/<P>/problem.toml} sets $n_{\min}$, the directions, the budget of gate 6 and the oracles, the constants that a record may not refer to: $f$ and the definitions it is built from.
\item \emph{Write the first Challenges.} \texttt{squeeze accept}, run once per direction, writes the statement of the seed and the first open Challenge, which asks to beat the seed.
\item \emph{Check the seeds.} \texttt{squeeze audit} checks that the seeds pass their own Challenges, with the three standard axioms only.
\item \emph{Describe the problem.} A \texttt{README.md} gives the statement, the known values with their sources, and the literature; \texttt{squeeze accept} later adds the record history to it.
\item \emph{Review.} A maintainer checks that the definition of $f$ denotes the intended object, including any reused Mathlib definition, and admits the problem. From then on it appears in \texttt{squeeze} like every other problem.
\end{enumerate}
The repository also contains a tutorial problem that is not part of the benchmark, the largest subset of $\{1,\dots,m\}$ that contains no element together with its double. Its $f$ is computable, so a newcomer can check values of $f$ by \texttt{decide} and practice the submission workflow on it.

\end{document}